\pdfoutput=1

\documentclass[11pt]{article}

\usepackage{coling}

\usepackage{times}
\usepackage{latexsym}

\usepackage[T1]{fontenc}

\usepackage[utf8]{inputenc}

\usepackage{microtype}

\usepackage{inconsolata}

\usepackage{graphicx}

\usepackage{amsmath} 
\usepackage{multicol} 
\usepackage{multirow} 
\usepackage{colortbl}
\usepackage{array}
\usepackage{booktabs}
\usepackage{longtable}
\usepackage{subcaption}
\usepackage{morefloats} 
\usepackage{etex} 
\usepackage{longtable}
\usepackage{makecell}

\title{Few Dimensions are Enough: Fine-tuning BERT with Selected Dimensions Revealed Its Redundant Nature}

\author{
  \textbf{Shion Fukuhata\textsuperscript{1}},
  \textbf{Yoshinobu Kano\textsuperscript{1}}
  \\
  \textsuperscript{1} Shizuoka University
  \\
  \small{
    \{sfukuhata, kano\}@kanolab.net
  }
}

\begin{document}
\maketitle
\begin{abstract}

When fine-tuning BERT models for specific tasks, it is common to select part of the final layer’s output and input it into a newly created fully connected layer. However, it remains unclear which part of the final layer should be selected and what information each dimension of the layers holds.
In this study, we comprehensively investigated the effectiveness and redundancy of token vectors, layers, and dimensions through BERT fine-tuning on GLUE tasks. The results showed that outputs other than the CLS vector in the final layer contain equivalent information, most tasks require only 2-3 dimensions, and while the contribution of lower layers decreases, there is little difference among higher layers.
We also evaluated the impact of freezing pre-trained layers and conducted cross-fine-tuning, where fine-tuning is applied sequentially to different tasks. The findings suggest that hidden layers may change significantly during fine-tuning, BERT has considerable redundancy, enabling it to handle multiple tasks simultaneously, and its number of dimensions may be excessive.

\end{abstract}

\section{Introduction} \label{sec:introduction}
Fine-tuning Transformer models \cite{vaswani2023attentionneed} for specific tasks often involves selecting outputs from the final layer and feeding them into a newly created fully connected  layer for task-specific learning \cite{devlin2019bertpretrainingdeepbidirectional} \cite{BERTlogy}.

By adding a special token [CLS] at the beginning of the input during pre-training, the final layer vector corresponding to this token (hereafter CLS vector) is commonly used to represent the entire sentence \cite{SentenceBERT} \cite{sun2020finetuneberttextclassification} \cite{BERTlogy}. However, it is unclear which part of the final layer is optimal for selection, or what information each dimension holds  \cite{clark2019doesbertlookat}. Additionally, using hidden layer outputs instead of the final layer has been shown to improve performance \cite{song2020utilizingbertintermediatelayers}, but the exact information encoded in each layer and dimension remains unknown \cite{tenney2019bertrediscoversclassicalnlp}.

By pre-training, models are expected to capture general linguistic representations \cite{kim2020pretrainedlanguagemodelsaware}. Fine-tuning leverages these pre-trained representations, but it is not well understood whether the newly added fully connected layer performs most of the learning if the hidden layers serve as fixed “language feature extractors,” \cite{howard2018universallanguagemodelfinetuning} \cite{irpan2019offpolicyevaluationoffpolicyclassification} or if catastrophic forgetting \cite{MCCLOSKEY1989109} causes significant changes in these layers \cite{li2024revisitingcatastrophicforgettinglarge}.

Several studies have reported that BERT contains excessive parameters and can be pruned without affecting performance \cite{DistilBERT}. The Lottery Ticket Hypothesis \cite{frankle2018lottery} suggests that within a dense, randomly initialized feedforward network, there exists a “winning ticket”—a sparse subnetwork that can achieve test accuracy comparable to the original network. This has been demonstrated in small networks like LeNet \cite{Lenet} for computer vision and is also applicable to LSTM \cite{LSTM}  and Transformers in NLP \cite{yu2020playinglotteryrewardsmultiple} \cite{chen2020lottery} \cite{prasanna2020bertplayslotterytickets}. In pre-trained BERT models, sparse subnetworks corresponding to 40\%-90\% of the tasks have been identified \cite{chen2020lottery}. However, while pruning reduces model parameters, it remains unclear which subnetworks are effective or what information each dimension contains.

In this study, we comprehensively examine what happens during fine-tuning of pre-trained BERT models using the GLUE benchmark \cite{wang2019gluemultitaskbenchmarkanalysis}, conducting dimension-wise, token-wise, and layer-wise comparisons (Figure \ref{fig1}). Our contributions are as follows:

\begin{figure*}[!t] 
  \centering 
  \includegraphics[width=0.80\textwidth]{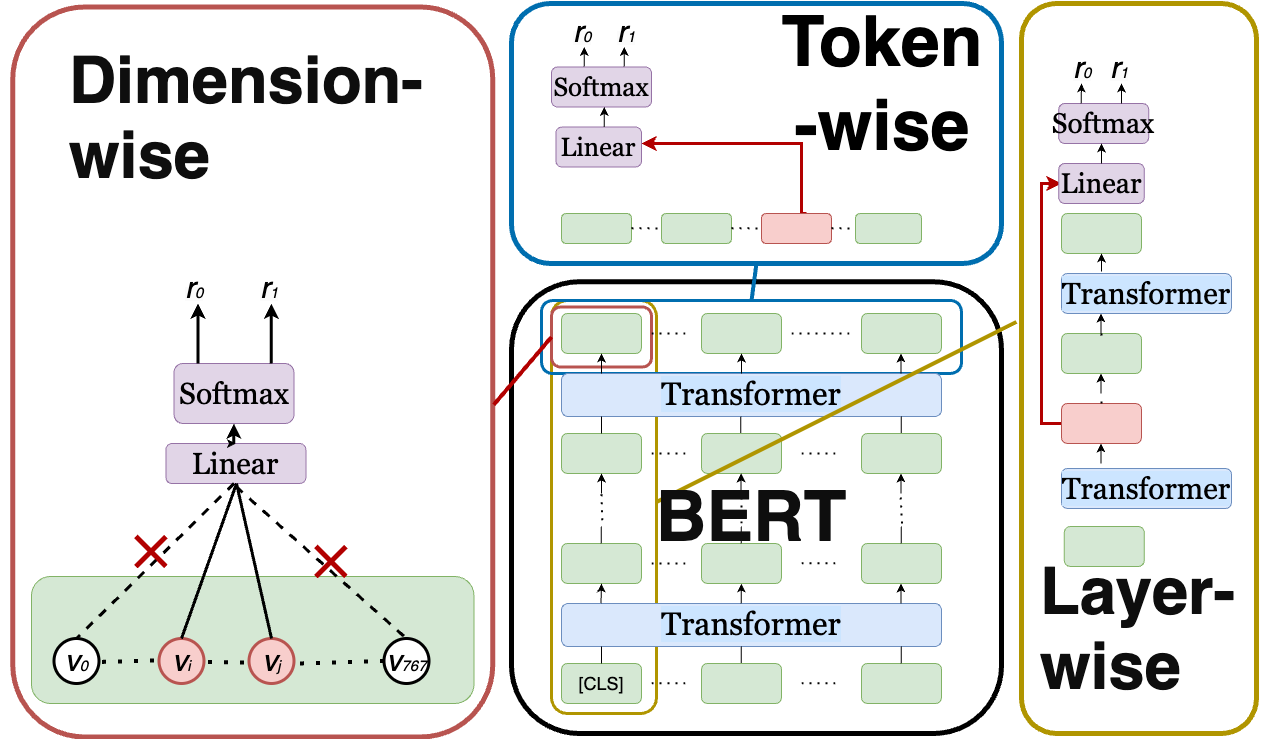} 
  \caption{
        Overview of the proposed method. 
        \textbf{Dimension-wise}: select a few dimensions,
        \textbf{Token-wise}: select a token vector by Max-Pooling,
        \textbf{Layer-wise}: select a few dimensions from hidden layers
    } 
  \label{fig1}
\end{figure*}

\begin{itemize}

\item We show that final layer vectors corresponding to tokens other than the CLS vector contain equivalent information (token-wise). 
\item We demonstrated that using only 2-3 dimensions from hidden layers reveals redundancy between layers and task-specific differences (layer-wise).
\item We showed that, despite significant changes during fine-tuning, the redundancy in hidden layers allows them to adapt to multiple tasks simultaneously (cross-fine-tune).
\end{itemize}

\section{Method} \label{sec:method}
The common assumption is that fine-tuning a pre-trained model enables it to perform various downstream classification tasks from the GLUE benchmark \cite{wang2019gluemultitaskbenchmarkanalysis}.

\paragraph{Token-wise Redundancy} \label{sec:method1}
We examine the redundancy of the CLS vector information by comparing inference using the other token vectors. Specifically, inference is performed using representations selected via MaxPooling based on norm size. 
If the CLS vector contains unique information, replacing it with MaxPooling-selected vectors should result in a significant performance drop.

\paragraph{Dimension-wise Redundancy} \label{sec:method2}
We investigate whether the vector used for downstream task inference is redundant and which dimensions are effective. 
Using DropConnect \cite{pmlr-v28-wan13}, we nullify subnetworks corresponding to unselected dimensions and infer using arbitrary two-dimensional combinations from the CLS vector.

\paragraph{Identifying Effective Dimensions} \label{sec:method3}
We assess the influence of individual dimensions on performance. First, we perform inference using three dimensions, then remove one at a time to observe the performance drop. 
If removing a specific dimension significantly reduces performance, it is likely important for the task. 
We then combine the identified dimensions to conduct inference with two dimensions. 
If all combinations yield high performance, the dimensions are considered effective for the task.

\paragraph{Layer-wise Redundancy} \label{sec:method4}
Similary, we evaluate redundancy in hidden layers.  
We select two dimensions from the CLS vector in the N-th layer and input only these into the final fully connected layer. 
If high performance is achieved without using the final layer, it indicates redundancy in higher layers. 
A significant performance drop at a specific hidden layer suggests that information up to that layer is sufficient for the task.

\paragraph{Freezing Pretrained Layers}
We compare performances between when the pre-trained layers are frozen with two fully connected layers added for fine-tuning, and when the layers are not frozen with one fully connected layer added for fine-tuning.

\paragraph{Cross-Fine-Tuning} \label{sec:method5}
If redundant, effective dimensions exist across tasks, a model fine-tuned on one task can be further fine-tuned on another without catastrophic forgetting. 
To test this, we perform cross-fine-tuning, evaluating the model’s performance after fine-tuning on one downstream task and then further fine-tuning on another. 
If the hypothesis holds, the performance will not degrade compared to directly fine-tuning on the second task.

\section{Experiment} \label{sec:experiment}
We used bert-base-uncased \cite{DBLP:journals/corr/abs-1810-04805} \footnote{https://huggingface.co/google-bert/bert-base-uncased} as the standard pre-trained model.
The downstream tasks were from the GLUE benchmark: CoLA, SST-2, MRPC, STS-B, QQP, MNLI, QNLI, and RTE.
For error analysis, only public training and validation data were used, split into an 8:1:1 ratio for training, validation, and testing.
During training, the CLS vector (or the MaxPooling-selected vector) of the pre-trained final layer was fed into a newly added fully connected layer. 
During inference on test data, we select dimensions and layers to fed into the fully connected layer.
Detailed settings described in Appendix \ref{sec:appendix:experiment_settings}.

\begin{table*}[ht]
  \centering
  \resizebox{0.85\textwidth}{!}{
  \begin{tabular}{|c|cc|cc|cc|cc|cc|cc|cc|cc|}
    \hline
    \textbf{Dimension Set} & \multicolumn{2}{c|}{\makecell{\textbf{MNLI}\\\textbf{Accuracy}}} & \multicolumn{2}{c|}{\makecell{\textbf{QQP}\\\textbf{Accuracy}}} & \multicolumn{2}{c|}{\makecell{\textbf{QNLI}\\\textbf{Accuracy}}} & \multicolumn{2}{c|}{\makecell{\textbf{STS-B}\\\textbf{Pcc}}} & \multicolumn{2}{c|}{\makecell{\textbf{MRPC}\\\textbf{Accuracy}}} & \multicolumn{2}{c|}{\makecell{\textbf{RTE}\\\textbf{Accuracy}}} & \multicolumn{2}{c|}{\makecell{\textbf{SST-2}\\\textbf{Accuracy}}} & \multicolumn{2}{c|}{\makecell{\textbf{CoLA}\\\textbf{Mcc}}} \\ \hline
    \textbf{}& \textbf{valid} & \textbf{test} & \textbf{valid} & \textbf{test} & \textbf{valid} & \textbf{test} & \textbf{valid} & \textbf{test} & \textbf{valid} & \textbf{test} & \textbf{valid} & \textbf{test} & \textbf{valid} & \textbf{test} & \textbf{valid} & \textbf{test} \\ \hline\hline
    \textbf{All (baseline)} & N/A & 83.2 & N/A & 90.1 & N/A & 88.7 & N/A & 88.8 & N/A & 77.9 & N/A & 67.5 & N/A & 94.8 & N/A & 53.1 \\ \hline\hline
    Set 1 & 75.1 & 75.1 & 89.1 & 89.0 & 88.5 & 87.9 & 83.2 & 86.1 & 78.4 & 76.5 & 70.4 & 65.3 & 94.8 & 94.2 & 60.3 & 54.3 \\ \hline
    Set 2 & 73.4 & 72.5 & 88.6 & 88.7 & 88.1 & 87.7 & 82.7 & 86.0 & 78.2 & 77.2 & 66.8 & 61.4 & 94.7 & 94.4 & 58.1 & 54.0 \\ \hline
    Set 3 & 73.1 & 73.2 & 88.6 & 88.6 & 88.1 & 87.4 & 82.6 & 85.8 & 77.7 & 78.9 & 66.4 & 62.8 & 94.6 & 94.8 & 57.7 & 55.3 \\ \hline
    Set 4 & 73.0 & 72.9 & 88.4 & 88.4 & 88.0 & 87.2 & 82.6 & 85.7 & 77.7 & 73.8 & 66.1 & 64.6 & 94.6 & 94.2 & 57.6 & 51.1 \\ \hline
    Set 5 & 72.8 & 72.2 & 88.1 & 88.0 & 87.9 & 87.3 & 82.6 & 85.8 & 77.7 & 77.7 & 66.1 & 66.1 & 94.6 & 94.6 & 57.6 & 53.9 \\ \hline
  \end{tabular}
  }
  \caption{Evaluation scores on \textbf{valid}ation and \textbf{test} sets for top 5 sets using two dimensions and \textbf{All} dimensions, showing comparable results by only two dimensions. (Pcc: Pearson Correlation Coefficient, Mcc: Matthews Correlation Coefficient)}
  \label{tab:2dims_Inference}
\end{table*}

\subsection{Token-wise: MaxPooling Output} \label{sec:experiment1}
Inference was performed using the token vector with the largest norm, selected via MaxPooling, compared with the CLS vector.

\subsection{Dimension-wise} \label{sec:experiment2}

\paragraph{Inference with Few Dimensions}
We evaluated whether inference using only a few dimensions from the final layer’s CLS vector could achieve sufficient accuracy. 
For each task, inference was repeated using randomly selected 2 or 3 dimensions. 
Multiple random sampling rates were tested when using 3 dimensions to assess their impact on performance, as described in \ref{sec:appendix:sampling_rate}.

\paragraph{Identifying Effective Dimensions} \label{sec:experiment3}
We investigated which dimensions were effective for each task. 
For the top 10 performing three-dimensional sets, we evaluated the effect of removing one dimension at a time on performance. 
Dimensions causing significant performance degradation upon removal were deemed effective. 
The top 5 effective dimensions were then combined for two-dimensional inference and evaluated.

\paragraph{Impact of Dropout} \label{sec:experiment6}
We investigated whether applying Dropout \cite{Dropout} during training affects the dimension-wise redundancy.

\subsection{Inference using Hidden Layers} \label{sec:experiment4}
We examined whether sufficient inference could be achieved using CLS vectors from layers other than the final layer. 
Two randomly selected dimensions from the CLS vector of each hidden layer were used. 
The focus was on whether the inference accuracy using lower-layer outputs was comparable to that of the final layer. 
The sampled dimension sets were consistent across all layers.

\subsection{Freezing Pretrained Layers}
We compared performances when freezing and not freezing pretrained layers.

\subsection{Cross-Task Fine-Tuning} \label{sec:experiment5}
We explored the impact of redundant dimensions. First, models were fine-tuned on one downstream task and then further fine-tuned on a different task. Performance was compared to a baseline where models were fine-tuned on each task individually. 
After resetting the fully connected layer, the model was fine-tuned again on each task, and performance differences were measured.

\section{Results}
\subsection{Token-wise: Output Selection by MaxPooling} \label{sec:results1}

Table \ref{tab:cls_vs_maxpooling} shows a comparison of the performance using the final layer’s CLS vector versus the MaxPooling-selected vector. In RTE, performance dropped when using the MaxPooling-selected vector, but no performance drop in other tasks.

\subsection{Dimension-wise} \label{sec:results2}
\paragraph{Inference with Few Dimensions}
The performance of the top five dimension sets, which achieved the highest accuracy on the validation data, is shown in Table \ref{tab:2dims_Inference}, \ref{tab:3dims_Inference}.
For all tasks except MNLI, the performance on the test data using the best-performing dimension set was nearly equivalent to using all dimensions. Table \ref{tab_Random_Seeds} shows the results of using five different seed values for inference with all dimensions, which helped estimate the performance equivalence range.

\begin{table*}[ht]
  \small
  \centering
  \resizebox{0.90\textwidth}{!}{
    \begin{tabular}{|c|cc|cc|cc|cc|cc|cc|cc|cc|}
      \hline
      \textbf{Seed} & \multicolumn{2}{c|}{\makecell{\textbf{MNLI}\\\textbf{Acc}}} & \multicolumn{2}{c|}{\makecell{\textbf{QQP}\\\textbf{Acc}}} & \multicolumn{2}{c|}{\makecell{\textbf{QNLI}\\\textbf{Acc}}} & \multicolumn{2}{c|}{\makecell{\textbf{STS-B}\\\textbf{Corr}}} & \multicolumn{2}{c|}{\makecell{\textbf{MRPC}\\\textbf{Acc}}} & \multicolumn{2}{c|}{\makecell{\textbf{RTE}\\\textbf{Acc}}} & \multicolumn{2}{c|}{\makecell{\textbf{SST-2}\\\textbf{Acc}}} & \multicolumn{2}{c|}{\makecell{\textbf{CoLA}\\\textbf{Mcc}}} \\ \hline
      & \textbf{valid} & \textbf{test} & \textbf{valid} & \textbf{test} & \textbf{valid} & \textbf{test} & \textbf{valid} & \textbf{test} & \textbf{valid} & \textbf{test} & \textbf{valid} & \textbf{test} & \textbf{valid} & \textbf{test} & \textbf{valid} & \textbf{test} \\ \hline
      \begin{tabular}[c]{@{}c@{}}42\end{tabular} & 83.5 & 83.2 & 90.1 & 90.1 & 89.6 & 88.7 & 85.8 & 88.7 & 77.5 & 77.9 & 65.7 & 67.5 & 95.2 & 94.8 & 55.8 & 53.1 \\ \hline
      \begin{tabular}[c]{@{}c@{}}0\end{tabular} & 83.2 & 83.1 & 90.2 & 90.0 & 89.5 & 89.0 & 84.2 & 88.2 & 79.9 & 79.7 & 61.7 & 65.7 & 95.1 & 95.0 & 56.3 & 50.3 \\ \hline
      \begin{tabular}[c]{@{}c@{}}331\end{tabular} & 83.1 & 83.1 & 90.2 & 90.0 & 89.5 & 89.0 & 85.6 & 88.7 & 78.4 & 78.2 & 65.7 & 67.2 & 94.3 & 94.3 & 56.7 & 54.0 \\ \hline
      \begin{tabular}[c]{@{}c@{}}17\end{tabular} & 83.2 & 83.3 & 90.1 & 90.0 & 89.5 & 88.7 & 84.8 & 88.5 & 79.2 & 80.4 & 66.8 & 67.2 & 94.5 & 94.9 & 58.0 & 54.3 \\ \hline
      \begin{tabular}[c]{@{}c@{}}31\end{tabular} & 83.3 & 83.2 & 90.0 & 90.0 & 89.5 & 88.8 & 83.9 & 88.3 & 78.9 & 78.2 & 66.8 & 64.6 & 95.0 & 94.9 & 56.4 & 54.5 \\ \hline
    \end{tabular}
  }
  \caption[Inference evaluation scores for different random seed initialization values]{Inference performance on each task for different seed values.}
  \label{tab_Random_Seeds}
\end{table*}

\begin{table*}[ht]
  \small
  \centering
  \begin{subtable}[t]{0.66\textwidth}
    \centering
    \resizebox{\textwidth}{!}{
    \begin{tabular}{|c|cc|cc|cc|cc|cc|}
      \hline
      & \multicolumn{2}{c|}{\makecell{\textbf{STS-B}\\Corr}} & \multicolumn{2}{c|}{\makecell{\textbf{MRPC}\\Acc}} & \multicolumn{2}{c|}{\makecell{\textbf{RTE}\\Acc}} & \multicolumn{2}{c|}{\makecell{\textbf{SST-2}\\Acc}} & \multicolumn{2}{c|}{\makecell{\textbf{CoLA}\\Mcc}} \\ \hline
      \textbf{Rank} & \textbf{valid} & \textbf{test} & \textbf{valid} & \textbf{test} & \textbf{valid} & \textbf{test} & \textbf{valid} & \textbf{test} & \textbf{valid} & \textbf{test} \\ \hline\hline
      All (baseline) & N/A & 88.8 & N/A & 77.9 & N/A & 67.5 & N/A & 94.8 & N/A & 53.1 \\ \hline\hline
      1 & 83.7 & 86.5 & 80.9 & 76.7 & 68.6 & 69.0 & 94.9 & 94.7 & 61.2 & 58.5 \\ \hline
      2 & 83.6 & 86.9 & 79.7 & 77.5 & 68.2 & 64.3 & 94.7 & 94.3 & 60.9 & 58.5 \\ \hline
      3 & 83.5 & 86.3 & 79.2 & 78.9 & 68.2 & 61.4 & 94.6 & 94.5 & 60.2 & 58.9 \\ \hline
      4 & 83.5 & 86.5 & 79.2 & 79.2 & 68.2 & 63.2 & 94.6 & 94.3 & 60.1 & 54.4 \\ \hline
      5 & 83.5 & 86.4 & 79.2 & 74.0 & 67.9 & 62.8 & 94.6 & 94.4 & 59.3 & 55.6 \\ \hline
    \end{tabular}
    }
    \caption{sampling: 75,202}
    \label{tab_3dims_Inference_75202}
  \end{subtable}
  \begin{subtable}[t]{0.66\textwidth}
    \centering
    \resizebox{\textwidth}{!}{
    \begin{tabular}{|c|cc|cc|cc|cc|cc|}
      \hline
      & \multicolumn{2}{c|}{\makecell{\textbf{STS-B}\\Corr}} & \multicolumn{2}{c|}{\makecell{\textbf{MRPC}\\Acc}} & \multicolumn{2}{c|}{\makecell{\textbf{RTE}\\Acc}} & \multicolumn{2}{c|}{\makecell{\textbf{SST-2}\\Acc}} & \multicolumn{2}{c|}{\makecell{\textbf{CoLA}\\Mcc}} \\ \hline
      \textbf{Rank} & \textbf{valid} & \textbf{test} & \textbf{valid} & \textbf{test} & \textbf{valid} & \textbf{test} & \textbf{valid} & \textbf{test} & \textbf{valid} & \textbf{test} \\ \hline\hline
      All (baseline) & N/A & 88.8 & N/A & 77.9 & N/A & 67.5 & N/A & 94.8 & N/A & 53.1 \\ \hline\hline
      1 & 83.5 & 86.3 & 79.7 & 77.5 & 67.5 & 62.1 & 94.9 & 94.7 & 60.1 & 54.4 \\ \hline
      2 & 83.3 & 86.3 & 78.9 & 77.9 & 67.1 & 65.7 & 94.7 & 94.3 & 58.9 & 56.8 \\ \hline
      3 & 82.9 & 85.5 & 78.2 & 77.9 & 66.8 & 61.0 & 94.6 & 94.5 & 58.7 & 52.3 \\ \hline
      4 & 82.9 & 86.3 & 77.7 & 75.5 & 66.4 & 63.9 & 94.6 & 94.3 & 58.1 & 52.6 \\ \hline
      5 & 82.9 & 85.5 & 77.7 & 77.5 & 66.4 & 58.8 & 94.6 & 94.4 & 58.1 & 53.5 \\ \hline
    \end{tabular}
    }
    \caption{sampling: 7,520}
    \label{tab:3dims_Inference_7520}
  \end{subtable}
  \begin{subtable}[t]{\textwidth}
    \centering
    \resizebox{\textwidth}{!}{
    \begin{tabular}{|c|cc|cc|cc|cc|cc|cc|cc|cc|}
      \hline
      & \multicolumn{2}{c|}{\makecell{\textbf{MNLI}\\Acc}} & \multicolumn{2}{c|}{\makecell{\textbf{QQP}\\Acc}} & \multicolumn{2}{c|}{\makecell{\textbf{QNLI}\\Acc}} & \multicolumn{2}{c|}{\makecell{\textbf{STS-B}\\Corr}} & \multicolumn{2}{c|}{\makecell{\textbf{MRPC}\\Acc}} & \multicolumn{2}{c|}{\makecell{\textbf{RTE}\\Acc}} & \multicolumn{2}{c|}{\makecell{\textbf{SST-2}\\Acc}} & \multicolumn{2}{c|}{\makecell{\textbf{CoLA}\\Mcc}} \\ \hline
      \textbf{Rank} & \textbf{valid} & \textbf{test} & \textbf{valid} & \textbf{test} & \textbf{valid} & \textbf{test} & \textbf{valid} & \textbf{test} & \textbf{valid} & \textbf{test} & \textbf{valid} & \textbf{test} & \textbf{valid} & \textbf{test} & \textbf{valid} & \textbf{test} \\ \hline\hline
      \textbf{All (baseline)} & N/A & 83.2 & N/A & 90.1 & N/A & 88.7 & N/A & 88.8 & N/A & 77.9 & N/A & 67.5 & N/A & 94.8 & N/A & 53.1 \\ \hline\hline
      1 & 79.3 & 78.81 & 88.9 & 88.7 & 88.9 & 88.3 & 82.6 & 85.9 & 76.7 & 75.7 & 66.4 & 63.9 & 94.4 & 94.3 & 56.3 & 50.8 \\ \hline
      2 & 77.6 & 77.0 & 88.6 & 88.5 & 88.6 & 87.8 & 82.5 & 84.5 & 76.5 & 77.0 & 66.1 & 60.6 & 94.4 & 94.2 & 56.2 & 52.4 \\ \hline
      3 & 77.1 & 76.8 & 88.3 & 87.9 & 88.1 & 87.7 & 81.9 & 85.2 & 76.2 & 75.0 & 65.3 & 59.6 & 94.4 & 94.3 & 56.2 & 52.4 \\ \hline
      4 & 76.04 & 75.8 & 88.3 & 87.9 & 87.8 & 86.9 & 80.8 & 83.4 & 76.0 & 78.9 & 64.6 & 62.1 & 94.3 & 94.2 & 55.9 & 49.4 \\ \hline
      5 & 75.5 & 75.3 & 88.2 & 88.4 & 87.6 & 87.0 & 80.1 & 83.4 & 76.0 & 75.0 & 64.6 & 62.1 & 94.2 & 93.5 & 55.4 & 51.9 \\ \hline
    \end{tabular}
    }
    \caption{sampling: 752}
    \label{tab:3dims_Inference_752}
  \end{subtable}
  \begin{subtable}[t]{\textwidth}
    \centering
    \resizebox{\textwidth}{!}{
    \begin{tabular}{|c|cc|cc|cc|cc|cc|cc|cc|cc|}
      \hline
      & \multicolumn{2}{c|}{\makecell{\textbf{MNLI}\\Acc}} & \multicolumn{2}{c|}{\makecell{\textbf{QQP}\\Acc}} & \multicolumn{2}{c|}{\makecell{\textbf{QNLI}\\Acc}} & \multicolumn{2}{c|}{\makecell{\textbf{STS-B}\\Corr}} & \multicolumn{2}{c|}{\makecell{\textbf{MRPC}\\Acc}} & \multicolumn{2}{c|}{\makecell{\textbf{RTE}\\Acc}} & \multicolumn{2}{c|}{\makecell{\textbf{SST-2}\\Acc}} & \multicolumn{2}{c|}{\makecell{\textbf{CoLA}\\Mcc}} \\ \hline
      \textbf{Rank} & \textbf{valid} & \textbf{test} & \textbf{valid} & \textbf{test} & \textbf{valid} & \textbf{test} & \textbf{valid} & \textbf{test} & \textbf{valid} & \textbf{test} & \textbf{valid} & \textbf{test} & \textbf{valid} & \textbf{test} & \textbf{valid} & \textbf{test} \\ \hline\hline
      \textbf{All (baseline)} & N/A & 83.2 & N/A & 90.1 & N/A & 88.7 & N/A & 88.8 & N/A & 77.9 & N/A & 67.5 & N/A & 94.8 & N/A & 53.1 \\ \hline\hline
      1 & 71.7 & 71.2 & 87.8 & 87.5 & 88.1 & 87.7 & 79.7 & 83.6 & 75.23 & 77.23 & 64.6 & 62.1 & 94.4 & 94.3 & 51.2 & 49.5 \\ \hline
      2 & 70.3 & 70.3 & 87.5 & 87.3 & 87.1 & 86.7 & 79.6 & 80.8 & 75.0 & 76.2 & 61.0 & 60.3 & 94.3 & 94.2 & 50.2 & 46.8 \\ \hline
      3 & 70.2 & 70.0 & 87.4 & 87.2 & 86.7 & 86.2 & 78.5 & 81.7 & 74.3 & 73.3 & 60.7 & 56.0 & 94.2 & 93.7 & 50.0 & 43.7 \\ \hline
      4 & 69.2 & 68.9 & 87.3 & 86.7 & 86.6 & 86.2 & 78.0 & 80.6 & 73.8 & 73.0 & 60.7 & 62.1 & 93.3 & 93.2 & 45.7 & 46.9 \\ \hline
      5 & 66.8 & 66.8 & 87.3 & 87.1 & 85.6 & 84.9 & 75.6 & 79.1 & 73.3 & 75.0 & 60.7 & 58.1 & 93.0 & 92.9 & 45.6 & 42.0 \\ \hline
    \end{tabular}
    }
    \caption{sampling: 75}
    \label{tab:3dims_Inference_75}
  \end{subtable}
  \caption{Top five evaluation results on the validation dataset of inference with 3 dimensions, and corresponding evaluation scores for test dataset, for four different numbers of combination sampling.}
  \label{tab:3dims_Inference}
\end{table*}

\begin{figure*}[!t] 
  \centering 
  \includegraphics[width=0.95\linewidth]{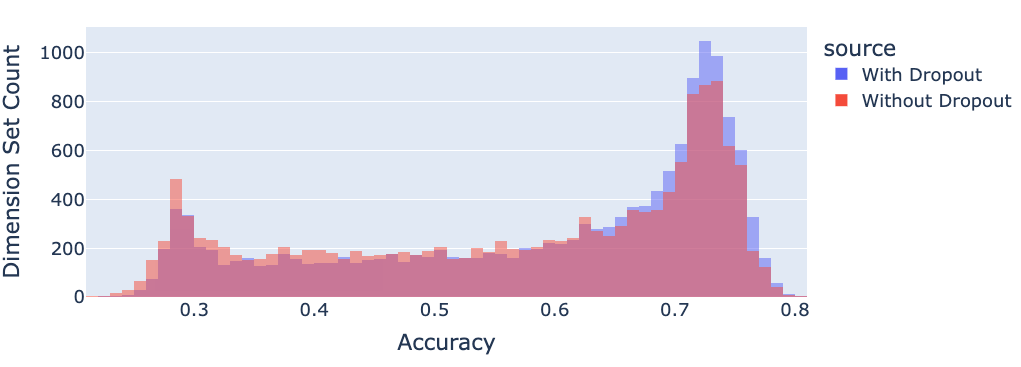} 
  \caption{
        The distribution of inference performance on the MRPC dataset using two dimensions, comparing cases with and without Dropout. The vertical axis represents the number of dimension combinations, and the horizontal axis represents inference performance.
    } 
  \label{fig:Dropout}
\end{figure*}

\paragraph{Identifying Effective Dimensions}
Table \ref{tab:effective_dimension_performance} shows the performance of the top five three-dimensional sets when one dimension was removed. Some dimensions caused a significant performance drop when removed. 
The top five dimensions with the largest drop were selected as effective, and the results of two-dimensional inference using these dimensions are shown in Table \ref{tab:combinations of Effective dimensions}. 
For all tasks except MNLI and CoLA, two-dimensional inference performed comparably to using all dimensions. As noted earlier, with three dimensions, performance was equivalent for all tasks except MNLI, suggesting that the number of effective dimensions for simpler tasks is generally 2 or 3.

\paragraph{Impact of Dropout}
With Dropout, there were more high-performing dimension combinations, but the difference is slight (Figure \ref{fig:Dropout}); Without Dropout, similar tendency of redundancy observed, thus Dropout should not be an essential factor of the redundancy.

\subsection{Inference Using Hidden Layers} \label{sec:results3}
Table \ref{tab:Inference performance when using CLS outputs below 11 layers} shows the results of inference using two-dimension from hidden layers, the top five dimension sets for each layer on the validation data. 
For QQP, MRPC, SST-2, STS-B, and CoLA, the performance of the best dimension set was similar between layers 12 and 11. However, for MNLI, QNLI, and RTE, there was a significant performance drop when using layer 11 instead of layer 12. Additionally, some tasks showed a sharp performance drop beyond a specific layer. 
For example, STS-B between layers 5-4, SST-2 between layers 10-9 and 9-8, and CoLA between layers 10-9.

\begin{table}[t]
  \small
  \centering
  \resizebox{0.48\textwidth}{!}{
  \begin{tabular}{|c|c|c|c|c|c|c|c|c|}
    \hline
    & {\makecell{\textbf{MNLI}\\Acc}} & {\makecell{\textbf{QQP}\\Acc}} &{\makecell{\textbf{QNLI}\\Acc}} & {\makecell{\textbf{STS-B}\\Corr}} & {\makecell{\textbf{MRPC}\\Acc}} & {\makecell{\textbf{RTE}\\Acc}} & {\makecell{\textbf{SST-2}\\Acc}} & {\makecell{\textbf{CoLA}\\Mcc}} \\ \hline  
    \textbf{CLS} & 83.2 & 90.1 & 88.7 & 88.8 & 77.9 & 67.5 & 94.8 & 53.1 \\ \hline
    \textbf{MaxPooling} & 83.4 & 89.8 & 90.0 & 88.1 & 77.9 & 56.3 & 94.8 & 58.2 \\ \hline
  \end{tabular}
  }
  \caption{Performance comparison between CLS and MaxPooling.}
  \label{tab:cls_vs_maxpooling}
\end{table}

\subsection{Results of Inference Using Token Vectors Selected by MaxPooling}
\label{sec:result4}
Table \ref{tab:cls_vs_maxpooling} compares the performances between using the CLS vector and the token vectors selected by MaxPooling. 
In RTE, performance decreased when using the vector selected by MaxPooling, but in other tasks, no significant performance degradation was observed.


\subsection{Freezing Pretrained Layers}
Adding two fully connected layers resulted in a greater performance drop when the layers were frozen compared to adding one fully connected layer without freezing (Table \ref{tab:Freezing Pretrained Layers}). This suggests that the Transformer layers learn downstream tasks more effectively than the fully connected layers.

\begin{table}[ht]
  \small
  \centering
  \resizebox{0.48\textwidth}{!}{
  \begin{tabular}{|c|c|c|c|c|c|c|c|c|}
    \hline
    & {\makecell{\textbf{MNLI}\\Acc}} & {\makecell{\textbf{QQP}\\Acc}} &{\makecell{\textbf{QNLI}\\Acc}} & {\makecell{\textbf{STS-B}\\Corr}} & {\makecell{\textbf{MRPC}\\Acc}} & {\makecell{\textbf{RTE}\\Acc}} & {\makecell{\textbf{SST-2}\\Acc}} & {\makecell{\textbf{CoLA}\\Mcc}} \\ \hline  
    \textbf{Not Freezing} & 83.2 & 90.1 & 88.7 & 88.8 & 77.9 & 67.5 & 94.8 & 53.1 \\ \hline
    \textbf{Freezing} & 54.5 & 73.5 & 69.1 & 19.7 & 69.6 & 59.6 & 78.4 & 0.0 \\ \hline
  \end{tabular}
  }
  \caption{Performance comparison between freezing and not freezing pretrained layers}
  \label{tab:Freezing Pretrained Layers}
\end{table}

\subsection{Cross-Task Fine-Tuning}
Regardless of the task, cross-fine-tuning achieved inference accuracy comparable to direct fine-tuning (Table \ref{tab:Cross_Finetune}).

\begin{table*}[ht]
  \small
  \centering
    \begin{tabular}{|c|c|c|c|c|c|c|c|c|c|}
      \hline
      \multicolumn{2}{|c|}{} & \multicolumn{8}{c|}{\textbf{Target}} \\ \cline{3-10}
      \multicolumn{2}{|c|}{} & {\makecell{\textbf{MNLI}\\Acc}} & {\makecell{\textbf{QQP}\\Acc}} &{\makecell{\textbf{QNLI}\\Acc}} & {\makecell{\textbf{STS-B}\\Corr}} & {\makecell{\textbf{MRPC}\\Acc}} & {\makecell{\textbf{RTE}\\Acc}} & {\makecell{\textbf{SST-2}\\Acc}} & {\makecell{\textbf{CoLA}\\Mcc}}\\ \hline
      \multirow{8}{*}{\textbf{Source}} & \textbf{MNLI}  & \cellcolor{gray!30}83.2 & 88.9 &  90.0 &  88.2 & 79.9 & 68.6 & 95.1 & 54.6  \\ 
      \cline{2-10} & \textbf{QQP}  & 83.1 & \cellcolor{gray!30}88.7  & 89.9 & 87.2 & 78.9 & 65.3 & 94.9 & 54.4\\ 
      \cline{2-10} & \textbf{QNLI}   & 82.8 & 87.8 & \cellcolor{gray!30}90.1 & 88.2 & 81.1 & 67.1 & 94.9 & 50.6 \\ 
      \cline{2-10} & \textbf{STS-B}  & 82.8 & 88.9 & 90.4 & \cellcolor{gray!30}88.8 & 78.7 & 64.2 & 94.7 & 52.3 \\ 
      \cline{2-10} & \textbf{MRPC}    & 82.3 & 88.6 & 89.8 & 88.3 & \cellcolor{gray!30}77.9 & 65.0 & 95.1 & 53.4 \\ 
      \cline{2-10} & \textbf{RTE}    & 82.3 & 88.8 & 89.9 & 87.9 & 76.7 & \cellcolor{gray!30}67.5 & 95.0 & 53.4 \\ 
      \cline{2-10} & \textbf{SST-2} & 83.2 & 88.7 & 90.0 & 88.0 & 79.7 & 68.6 & \cellcolor{gray!30}94.8 & 54.7 \\ 
      \cline{2-10} & \textbf{CoLA}    & 83.1 & 89.0 & 90.0 & 87.8 & 81.6 & 63.2 & 95.1 & \cellcolor{gray!30}53.1 \\ \hline
    \end{tabular}
  \caption{Cross-fine-tuning results.
    Each cell shows the inference evaluation score on the target task, after fine-tuning the model on the source task followed by additional fine-tuning on the target task.
    The diagonal entries represent the inference scores when directly fine-tuning on the target task.}
  \label{tab:Cross_Finetune}
\end{table*}




\subsection{Redundancy in the Number of Dimensions} \label{sec:discussion1}
We confirmed that inference using only two dimensions could achieve comparable accuracy to using all dimensions, provided a good set of dimensions was selected. 
This was particularly evident in simpler tasks, such as sentiment classification in SST-2 and grammatical correctness judgment in CoLA, where accuracy was on par with using all dimensions. 
This suggests that sparse subnetworks corresponding to good dimension sets may contain as much information as dense networks for these tasks. 
In this regard, the number of dimensions used during model inference may be redundant, especially for simpler tasks.

\subsection{Effectiveness of Each Dimension} \label{sec:discussion2}
It was observed that certain dimensions, when removed, led to significant performance degradation, and using these dimensions resulted in high inference accuracy. 
This suggests that the dimensions effective for each task may vary, and that only specific sparse subnetworks are heavily influenced by task-specific learning.

\subsection{Redundancy in the Number of Transformer Layers} \label{sec:discussion3}
The number of Transformer layers contributed to performance. Notably, for SST-2 and CoLA, there was a significant drop in inference performance beyond certain layers. 
This indicates that the information up to those specific layers may be crucial for these tasks. 
In general, the redundancy in the number of Transformer layers appears to be minimal, but this may not hold true for certain tasks.

\subsection{Impact of Redundant Dimensions} \label{sec:discussion4}
The results of cross-finetuning revealed that the inference performance when a model trained on Task A was further fine-tuned on Task B showed no difference compared to directly training on Task B. 
Considering the redundancy in dimensions required for inference, it is possible that the dimensions learned for different tasks vary. The redundancy in dimensions in existing models may contribute to their adaptability to a wide range of tasks.

\section{Conclusion}


We showed that BERT exhibits redundancy across token vectors, dimensions, and layers, with performance maintained even when significantly removing the dimensions. 
Fine-tuning alters the Transformer layer weights but retains previously learned information due to this redundancy. These results suggest a potential for optimizing BERT models.

\section*{Limitations}



\paragraph{Sampling}
In this study, inference using only two or three dimensions was conducted with randomly sampled combinations. While some dimension combinations were not tested, the claim of this study is that selecting an optimal set of dimensions can achieve performance equivalent to using all dimensions. Therefore, the current experimental setup is deemed sufficient.

\paragraph{GPT Models}
This study focuses on BERT models. While similar redundancy may exist in GPT models, investigating this is left for future work.

\paragraph{Other Languages}
This study examined the English BERT model. Similar redundancy may exist in other languages, but verifying this is also a subject for future research.

\section*{Acknowledgments}
The present study was supported by the JSPS Kakenhi (JP22H00804), JST PRESTO (JPMJPR2461), JST AIP Accelerated Research Program (JPMJCR22U4), and the SECOM Science and Technology Foundation's Specific Area Research Grant.

\bibliography{custom}

\appendix

\section{Experimental Settings} \label{sec:appendix:experiment_settings}

\subsection{Experimental Setup for Fine-Tuning} \label{sec:appendix:experiment_settings}
Dropout rate: 0.1 \\
Learning rate: 5e-5 \\
Batch size: 64 \\
Maximum Training Epochs: 5 \\
Random seed: 42 \\

\subsection{Sampling Rate for Inference Using Two or Three Dimensions} \label{sec:appendix:sampling_rate}
Different sampling rates were used depending on the dataset, considering execution time. 

\subsubsection{Using Two Dimensions}
We have $_{768}C_2=294,528$ combinations of dimensions in total, which correspond to 100\% in the following values. For each task, we executed sampling rates of:
MLNI, QQP, QNLI:  1\% (2,945) \\
STS-B, MRPC, RTE, SST-2, CoLA: 5\% (14,726) \\

\subsubsection{Using Three Dimensions}
We have $_{768}C_3=75,202,816$ combinations of dimensions in total, which correspond to 100\% in the following values. For each task, we executed sampling rates of:
MLNI, QQP, QNLI: \\ 0.001\% (752) / 0.0001\% (75)\\
STS-B, MRPC, RTE, SST-2, CoLA: \\ 0.1\% (75,202) / 0.01\% (7,520) / 0.001\% (752) / 0.0001\% (75)\\

\section{Error Analysis}

We conducted an analysis of the cases in which fails when using two dimensions.
For the CoLA task, we investigated the data that was correctly classified when using all dimensions but was consistently misclassified by the top five high-performing dimension sets.
The results showed that the misclassified data included sentences with reversed word order (e.g., “this problem, the sooner you solve the more easily you’ll satisfy the folks up at corporate headquarters.”) and 
sentences where the subject was omitted (e.g., “robin will eat cabbage but she won’t ice cream.”). It was confirmed that these challenging examples, which were labeled as grammatically correct, were incorrectly inferred as grammatically incorrect.

\renewcommand{\thetable}{A.\arabic{table}}
\renewcommand{\thefigure}{A.\arabic{figure}}
\setcounter{table}{0} 
\setcounter{figure}{0} 

\begin{table*}[htbp]
  \small
  \centering

  \caption{The evaluation scores on the validation dataset of the top 10 three-dimension set, where one dimension is removed for each. Numbers in the columns of \textbf{Original Dimension Set} and \textbf{Dimension Set} are unique dimension IDs. Red highlighted cells show top five sets on their performance. }
  \label{tab:effective_dimension_performance}
\end{table*}

\begin{table*}[htbp]
  \centering
  %
  \caption{Inference performance for combinations of effective dimensions. Numbers in the column of \textbf{Dimension Set} are unique dimension IDs.}
  \label{tab:combinations of Effective dimensions}
\end{table*}

\begin{table*}[htbp]
  \small
  \centering
  %
      }
      \caption{CoLA}
    \end{subtable} \\
  \end{tabular}
  \caption{Inference performance when using CLS tokens of 11 hidden layers}
  \label{tab:Inference performance when using CLS outputs below 11 layers}
\end{table*}


\end{document}